\PassOptionsToPackage{table,x11names}{xcolor}
\documentclass{article}

\PassOptionsToPackage{numbers, compress}{natbib}
    \usepackage[preprint]{neurips_2025}

\usepackage{tcolorbox}

\usepackage[utf8]{inputenc} % allow utf-8 input
\usepackage[T1]{fontenc}    % use 8-bit T1 fonts
\usepackage{hyperref}       % hyperlinks
\usepackage{url}            % simple URL typesetting
\usepackage{booktabs}       % professional-quality tables
\usepackage{amsfonts}       % blackboard math symbols
\usepackage{nicefrac}       % compact symbols for 1/2, etc.
\usepackage{microtype}      % microtypography
\usepackage{xcolor}       % colors

\usepackage{hyperref}       % hyperlinks
\hypersetup{
colorlinks= true,
urlcolor = blue,
linkcolor = red,
citecolor = blue,
}

\usepackage{xspace}
\usepackage{graphicx}
\usepackage{multicol}
\usepackage{multirow}
\usepackage{makecell}
\PassOptionsToPackage{numbers}{natbib}
\usepackage{wrapfig}
\usepackage{subcaption}
\usepackage{enumitem}
\usepackage{titletoc}
\usepackage{amsmath}
\usepackage{amssymb}
\usepackage{booktabs}    % For nice table lines
\usepackage{longtable}   % Optional: for long tables that span pages
\usepackage{array}       % For better column formatting
\usepackage{pifont}

\usepackage{listings}
\setlist[itemize]{leftmargin=*}

\title{Automated PRO-CTCAE Symptom Selection based on Prior Adverse Event Profiles}

\author{%
\textbf{Francois Vandenhende$^{1}$ \quad Anna Georgiou$^{1}$ \quad Michalis Georgiou$^{1}$}\\
\textbf{Theodoros Psaras$^{1}$ \quad Ellie Karekla$^{1}$} \\[6pt]
$^1$ClinBAY Limited, Limassol, Cyprus\\
Correspondence: \texttt{francois@clinbay.com}\\
\url{https://app.clinbay.com/safeterm}
}

\PassOptionsToPackage{numbers}{natbib}

\begin{document}

\maketitle

%%==================================%%
%% Abstract
%%==================================%%
\begin{abstract}
	The PRO-CTCAE is an NCI-developed patient-reported outcome system for capturing symptomatic adverse events in oncology trials. It comprises a large library drawn from the CTCAE vocabulary, and item selection for a given trial is typically guided by expected toxicity profiles from prior data. Selecting too many PRO-CTCAE items can burden patients and reduce compliance, while too few may miss important safety signals. We present an automated method to select a minimal yet comprehensive PRO-CTCAE subset based on historical safety data. Each candidate PRO-CTCAE symptom term is first mapped to its corresponding MedDRA Preferred Terms (PTs), which are then encoded into Safeterm—a high-dimensional semantic space capturing clinical and contextual diversity in MedDRA terminology. We score each candidate PRO item for relevance to the historical list of adverse event PTs and combine relevance and incidence into a utility function. Spectral analysis is then applied to the combined utility and diversity matrix to identify an orthogonal set of medical concepts that balances relevance and diversity. Symptoms are rank-ordered by importance, and a cut-off is suggested based on the explained information. The tool is implemented as part of the Safeterm trial-safety app. We evaluate its performance using simulations and oncology case studies in which PRO-CTCAE was employed. This automated approach can streamline PRO-CTCAE design by leveraging MedDRA semantics and historical data, providing an objective and reproducible method to balance signal coverage against patient burden.
\end{abstract}


\section{Introduction}

The National Cancer Institute (NCI) Patient-Reported Outcomes version of the Common Terminology Criteria for Adverse Events (PRO-CTCAE) allows patients in cancer trials to directly report symptomatic Adverse Events (AEs) \citep{Basch2014, Smith2016}. While the standard CTCAE contains approximately 790 distinct AE items, only about 10\% correspond to symptoms suitable for self-report. The PRO-CTCAE library was developed by translating each of these 80 CTCAE symptoms into 1--3 plain-language items (attributes) such as frequency, severity, interference, or presence/absence. In total, the PRO-CTCAE item library contains about 124 items representing these toxicities. This comprehensive coverage ensures patient-reported data complement the clinician-graded CTCAE grades, as the two assessments may differ \cite{Xiao2013}. However, it poses a challenge: presenting all 124 items would be impractical, whereas selecting a subset of items for a trial must not omit important effects.

Clinicians and trial designers typically select PRO-CTCAE symptoms based on the expected symptomatic profile of the regimen and/or the disease population \citep{Kluetz2016}. This process is often guided by prior trials, preclinical data, or disease context, supported by consensus-building reviews \cite{Reeve2014, Christiansen2023_PROCTCAE}. 

However, this symptoms selection process is inherently subjective and manual. If too few items are chosen, rare or unexpected symptoms may be missed; if too many are chosen, patient completion rates and data quality may suffer. Conditional branching and smart electronic data capture forms have been implemented to reduce patient burden \citep{Basch2014}, but there is currently no automated, objective standard for choosing which symptom terms to include. 

In short, PRO-CTCAE item selection is a complex trade-off between \textit{coverage} (capturing all relevant AEs) and \textit{conciseness} (minimizing burden). An ideal selection would cover every symptom concept expected based on past data without redundant overlap.

To address this, we propose an automated algorithmic approach that leverages historical AE data to objectively guide PRO-CTCAE item selection. Historical AE lists (derived from past trials, literature, or advisory boards) are standardized as Medical Dictionary for Regulatory Activities (MedDRA) \citep{ich1999_meddra_structure} Preferred Terms (PTs), allowing us to use Safeterm, a knowledge-enhanced MedDRA embedding system \citep{vandenhende2025}, to represent symptom terms in a high-dimensional semantic space. We then apply a numerical algorithm, based on spectral decomposition of a weighted relevance-diversity matrix, to identify distinct, clinically relevant medical concepts.  

Our method is a non-iterative, single-step alternative to the Determinantal Point Processes (DPP) iterative algorithm \citep{Kulesza2012DeterminantalPP} for generating diverse, optimal subsets. Unlike DPP, which relies on iterative sampling and probability distributions to ensure diversity, our approach directly extracts an orthogonal set of concepts in a single computational step, providing deterministic and reproducible selection. This makes it faster and more straightforward to apply in practice, while maintaining a balance between relevance and diversity.  

We assess the performance of this method using simulated adverse event lists where PT for target symptoms are mixed with random AEs. We further validate the practical utility of the approach through a retrospective case study involving successive clinical trials in Multiple Myeloma. 

\section{Methods}
The automated PRO-CTCAE item selection pipeline is designed to create a minimal yet comprehensive subset of symptoms by leveraging the semantic relationship between PRO-CTCAE symptoms and historical AEs. This relationship is studied within MedDRA terminology.

\subsection{Initial Data Processing and Semantic Embedding}
\subsubsection{PRO-CTCAE to MedDRA Mapping}
We begin by mapping the full PRO-CTCAE symptom library to MedDRA version 28.1. This establishes a common semantic ground for assessing both candidate symptoms and historical AEs, enabling the use of Safeterm's semantic encoding capabilities. The mapping was performed by a trained specialist. Each PRO symptom was linked to one, or occasionally two, corresponding MedDRA PTs. The process prioritized identical or conceptually best-match PTs, leveraging the formal MedDRA Lowest Level Terms (LLTs) where necessary. The complete mapping of PRO-CTCAE (v5/22/25) is provided in Table~\ref{tab:meddra_map}, and it is used strictly for semantic encoding and item selection purposes. Crucially, the resulting MedDRA PTs are not intended to replace the original PRO-CTCAE items, which remain the gold standard for patient-reported symptom capture.

\subsubsection{Embedding with Safeterm}

\begin{figure}
	\centering
	\includegraphics[width=1.0\textwidth]{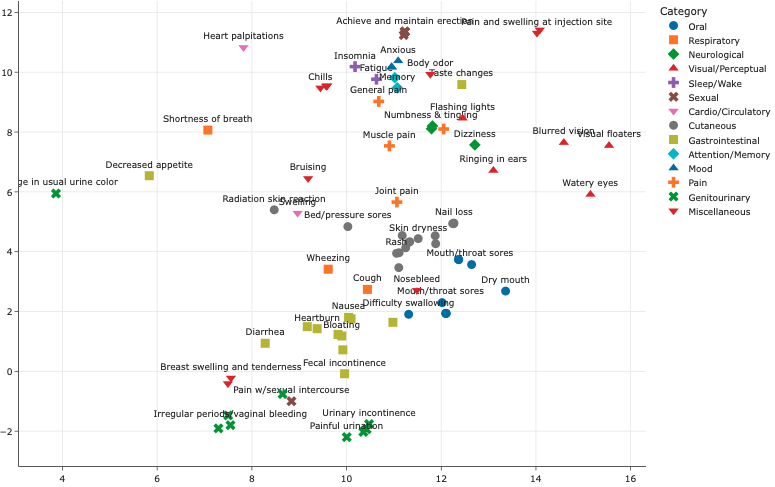}
	\caption{Safeterm 2-D map of PRO-CTCAE symptoms classified by category.}
	\label{fig_01}
\end{figure}

The core of our approach is the use of Safeterm \cite{vandenhende2025} to convert clinical terminology into high-dimensional numerical vectors known as embeddings. Safeterm is trained on a large medical text corpus and enriches the MedDRA terminology by adding data-driven semantic structure to its existing hierarchy. This results in a specialized embedding space in which every MedDRA PT is represented as a point whose location reflects its clinical meaning.
In this space, terms that are clinically similar lie close together, while dissimilar terms are farther apart \citep{Miller01011991}. By encoding both PRO-CTCAE symptoms and historical AE terms as vectors, we obtain a precise and quantitative way to measure their semantic similarity and inter-relationships. This embedded representation forms the foundation for scoring symptom relevance and identifying redundancy in the item-selection pipeline.
Figure~\ref{fig_01} presents the location of the 80 PRO-CTCAE symptoms into the Safeterm 2-D map. This map is a dimension reduction of the complete embedding space.
It shows how related PRO-CTCAE symptoms cluster together.

\subsection{Relevance/Redundancy Scoring}
\subsubsection{Symptom Redundancy}

To determine the semantic redundancy between symptoms, we quantify their similarity in the embedding space. To do so, each of the $N_{\text{PRO}}$ candidate items is mapped to its normalized Safeterm embedding, forming $\mathbf{E}_{\text{PRO}}$.
The semantic similarity matrix $\mathbf{S}$ (size $N_{\text{PRO}} \times N_{\text{PRO}}$) is computed using cosine similarity between all PRO embeddings.  
This matrix is essential for detecting pairs of symptoms that are semantically overlapping:
\[
\mathbf{S} = \mathbf{E}_{\text{PRO}} \cdot \mathbf{E}_{\text{PRO}}^T.
\]
$S_{i,j}$ quantifies how closely related symptoms $i$ and $j$ are in meaning, with values near 1 indicating high similarity and values near 0 indicating dissimilarity.

\subsubsection{PRO Relevance}

The relevance of each PRO item to the historical AE profile is quantified by comparing their embeddings. First, all historical AE PTs are mapped into Safeterm’s normalized embedding space, yielding the matrix $\mathbf{E}_{\text{trial}}$.

Each PRO item is then evaluated based on how closely its embedding aligns with the embeddings of the historical AEs. We compute a similarity matrix $\mathbf{Q}$ of size $N_{\text{trial}} \times N_{\text{PRO}}$ as
\[
\mathbf{Q} = \mathbf{E}_{\text{trial}} \cdot \mathbf{E}_{\text{PRO}}^{T},
\]
where each entry $\mathbf{Q}_{i,j}$ represents the semantic similarity between AE $i$ and PRO item $j$.

The relevance score for PRO item $j$ is defined as its strongest similarity to any historical AE:
\[
R_j = \max_i \, \mathbf{Q}_{i,j}.
\]
This score reflects how well each symptom corresponds to at least one AE in the historical profile.

\subsubsection{AE importance weights}

To account for the differing clinical importance or incidence of historical AEs, we introduce an AE-specific weighting scheme. Let $w_i > 0$ denote the weight assigned to AE $i$, which can be derived from its historical incidence, severity, or other prior knowledge. These weights reflect the relative importance of individual AEs in the context of the target population or mechanism of action.

We propagate AE weights to each PRO item by aggregating the weights of all AEs that are strongly semantically related to that item. Specifically, for PRO item $j$, we sum the weights of all AEs $i$ whose similarity score $Q_{ij}$ exceeds a high-similarity threshold, defined as
\[
Q_{ij} > \alpha \cdot \max_i (Q_{ij}),
\]
where we set $\alpha = 0.9$ to retain only the top 10\% of similarity relationships for each PRO item.

The resulting PRO-item weight is
\[
W_j = \sum_{i : Q_{ij} > \alpha \cdot \max_i(Q_{ij})} w_i,
\]
which quantifies the clinical importance of the PRO item based on the weighted contribution of closely related AEs.

\subsection{Utility Function}

For PRO item selection, we define a utility function that combines both the relevance ($\mathbf{R}$) and weight ($\mathbf{W}$) scores into a single
utility vector $\mathbf{U}$. It quantifies the overall importance of each PRO item.
The utility score $U_j$ is defined as:
\[
U_j = R^*_j + \beta \frac{W_j}{\max(W_j)}, 
\]
where

\begin{enumerate}
    \item $R^*$ is a logistic transform applied to $R_j$:
    \[
    R^*_j = \frac{1}{1 + e^{-k(R_j - x_0)}},
    \]
    which emphasizes the turning point between low and high relevance. With steepness $k = 20$ and midpoint $x_0 = 0.8$, similarity scores below 0.7 remain near zero ($R^* \approx 0$); the curve increases steeply between 0.7 and 0.9, and saturates above 0.9, treating very similar items as nearly equivalent.

    \item $\beta$ is a weighted tie-break. Using $\beta = 0.1$, the weight term provides a small additive boost to distinguish items with
    similar relevance. If no AE weights are supplied, $W_j$ is constant and this term may be omitted.
\end{enumerate}

\subsubsection{Combining Utility and Redundancy: The L-Kernel}

The L-Kernel \citep{Kulesza2012DeterminantalPP} integrates the item-level utility
vector $\mathbf{U}$ with the pairwise semantic similarity matrix $\mathbf{S}$:
\[
\mathbf{L}_{i,j} = U_i \cdot \mathbf{S}_{i,j} \cdot U_j.
\]

This construction combines two key components of the selection problem.  
First, the diagonal entries $\mathbf{L}_{j,j} = U_j^2$ encode the intrinsic importance of
each individual item. Second, the off-diagonal terms scale the similarity between pairs of
items by their respective utilities, allowing the matrix to represent both the quality of
items and the degree to which they overlap semantically. As a result, $\mathbf{L}$ provides a
joint measure of utility weighted by redundancy, forming the basis for identifying
subsets that are simultaneously informative and diverse.

\subsection{Symptoms Selection Based on Spectral Decomposition}

We select the number of PRO symptoms to retain and the individual items using a spectral analysis of the L-Kernel matrix $\mathbf{L}$. This technique is conceptually similar to Principal Component Analysis (PCA) \citep{Jolliffe2016}. 
It leverages the dimension reduction principle to focus selection on an orthogonal high-relevance subspace, capturing the key semantic variability among candidate symptoms. 

\subsubsection{Eigen-Decomposition of the L-Kernel}

The L-Kernel $\mathbf{L}$ is decomposed into its eigenvalues $\lambda_j$ and eigenvectors $\mathbf{V}$:

\[
\mathbf{L} = \mathbf{V} \, \text{diag}(\lambda_j) \, \mathbf{V}^T.
\]

The eigenvectors $\mathbf{V}_j$ represent orthogonal axes in the symptom relevance/diversity space, while the corresponding eigenvalues $\lambda_j$ quantify the variance captured along each axis.
Axes associated with larger eigenvalues carry more utility in the semantic space.

\subsubsection{Determination of Optimal Subspace Size}

To identify the number of meaningful axes, we compute the cumulative variance explained by the eigenvalues. The optimal number of axes, $k_{\text{optimal}}$, is defined as the smallest number of eigenvectors required to explain a target fraction of the total variance (default: 90\% to 97.5\%):

\[
k_{\text{optimal}} = 
\min \left\{
j \,\bigg|\, 
\frac{\sum_{i=1}^{j} \lambda_{\text{desc}, i}}
{\sum_{i=1}^{N} \lambda_i}
\geq \text{info}
\right\}.
\]

Because the axes are orthogonal, we hypothesize that each corresponds to a unique, relevant medical concept combining symptoms. 
We therefore suggest selecting, at a \textit{minimum}, a number of PRO-CTCAE symptoms equal to $k_{\text{optimal}}$. 
The actual number of retained symptoms may be larger, depending on their diversity leverage scores (see below).

\subsubsection{Ranking by Diversity Leverage}

Once $k_{\text{optimal}}$ is determined, individual PRO items are ranked according to their contribution to the top diversity axes. This approach directly measures each item's diversity leverage.

For symptom $j$, the diversity leverage score is defined as:

\[
\text{Leverage}_j = \sum_{i=1}^{k_{\text{optimal}}} 
\left(\mathbf{V}_{\text{top}, i, j}\right)^2.
\]

This sum captures the item's unique contribution across the most relevant $k_{\text{optimal}}$ diversity axes. High leverage values indicate that the symptom strongly defines multiple dimensions of the relevance space and is less redundant with other items.

Items are then ranked in descending order of diversity leverage. A minimum of $k_{\text{optimal}}$ items are selected to constitute the final minimal yet comprehensive set of PRO symptoms. 

\subsection{Implementation and Application}

We implemented this pipeline in Python and integrated it into the Safeterm application as an Application Programming Interface (API). Users can input historical AE data (as a list of MedDRA PTs, with optional incidence counts) and receive a ranked PRO-CTCAE item list, along with the recommended number of items to select. The output (see illustrative Table \ref{tab:pro_utility_table}) includes the ranked PRO items, their relevance, weight, and diversity scores, as well as the list of related AEs with relevance scores and incidence frequencies. A flag identifies items included in the $k_{\text{optimal}}$ deduplicated subset.

We assessed the method's performance using trial simulations and a real-world retrospective case study.

\subsubsection{Trial Simulations}

To assess the robustness of the PRO-CTCAE selection pipeline to noise and variable input profiles, we employed a large-scale Monte Carlo simulation ($N=100,000$).

The study design involved simulating trial AEs for known PRO-CTCAE symptoms (ranging from 5 to 40) with random noise terms (ranging from 10 to 50). Between 1 and 3 AEs were generated per symptom. The list of candidate terms was determined using the Safeterm Automated Medical Query (AMQ) capability. The AMQ takes a medical concept (e.g., a disease or symptom) as input and returns a list of associated MedDRA PTs with a relevance score above 75\%, thereby simulating a comprehensive expected toxicity profile. The random noise set consisted of all MedDRA PTs (approximately 27,000 terms), excluding any term from the AMQ set.

We used a variance threshold of $\text{info} = 97.5\%$ in the simulations and reported the number of True Positives (TP), False Positives (FP), and False Negatives (FN). Precision was derived as $\text{TP}/(\text{TP}+\text{FP})$,  recall as $\text{TP}/(\text{TP}+\text{FN})$ and F1 as the harmonic mean of precision and recall.

To determine the TP rate by symptom, we replicated the simulations with one symptom at a time. The overall performance was summarized using the mean, median, standard deviation, minimum, and maximum across simulations.

\subsubsection{Clinical Study}
We applied the system to two successive oncology trials of GSK2857916 in patients with multiple myeloma to evaluate its practical and predictive utility:
\begin{enumerate}
    \item \textbf{NCT02064387 (Phase I):} A dose escalation/expansion study. We utilized the full AE profile from the expansion multiple myeloma set (35 patients treated at the recommended phase II dose of 3.4 mg/kg) to serve as the historical AE incidence data for our selection tool.
    \item \textbf{NCT03525678 (Phase II):} An open-label, two-arm study in multiple myeloma, which utilized a pre-defined PRO-CTCAE instrument of 15 symptoms.
\end{enumerate}
We compared the PRO-CTCAE symptoms selected by our method (info = 90\%), based on the AE profile of NCT02064387, to the actual 15 items used in the subsequent NCT03525678 study. We also compared the performance of our PRO list to the pre-defined set in terms of coverage of observed AEs in NCT03525678.

\section{Results}

\subsection{Simulation Study}

Performance results from our simulation study are summarized in Table~\ref{tab:simulation_summary}. 
When simulating AEs related to between 5 and 40 PRO-CTCAE symptoms (mean = 22.5), the mean recall, precision, and F1 were all above 70\%. This demonstrates that the method can effectively recover relevant symptoms from noisy AE profiles while limiting the PRO-CTCAE instrument to the most relevant set.
Performance was stable across varying numbers of simulated symptoms and noise levels, indicating robustness to input variability. It was somewhat affected by the selected variance threshold; a lower value decreased recall but increased precision, as expected (results not shown). We therefore recommend careful calibration of this input parameter, choosing a larger percentile when the number of distinct AEs is higher.

\begin{figure}[ht]
    \centering
    \includegraphics[width=0.9\textwidth]{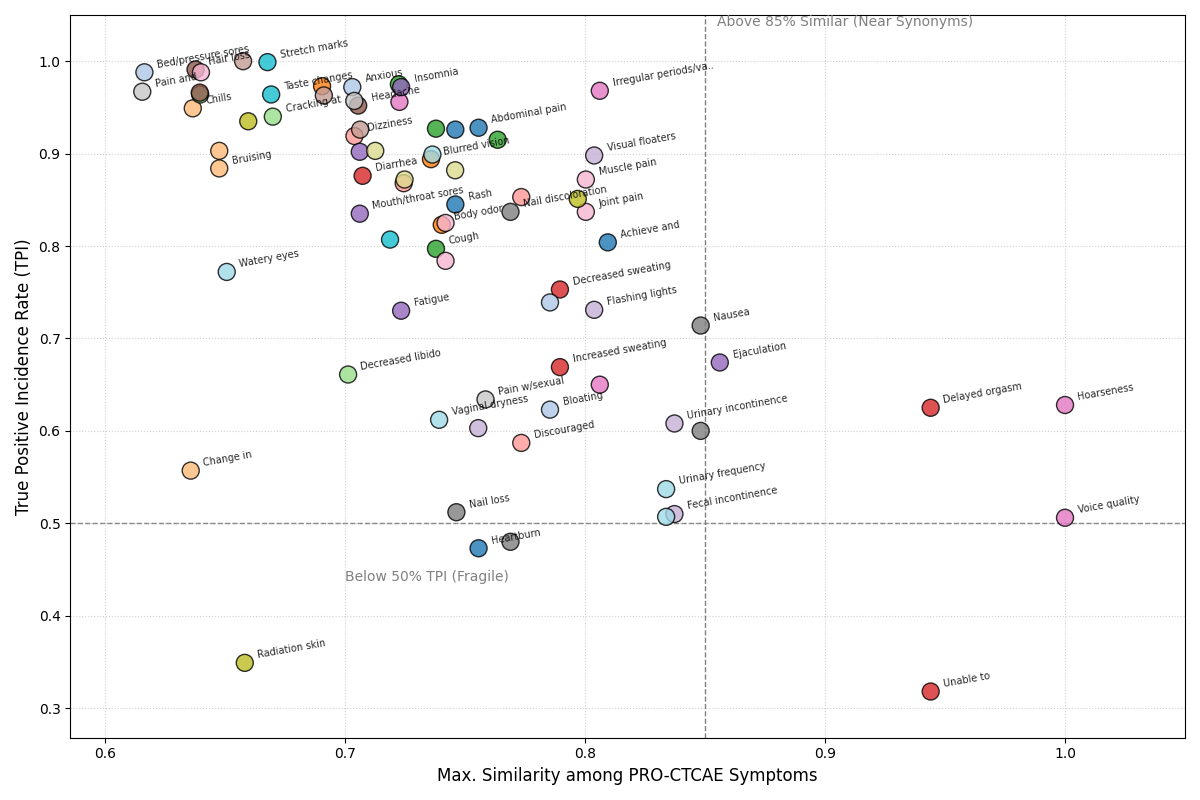}
    \caption{Maximum similarity among 80 PRO-CTCAE symptoms versus True Positive Incidence Rate (Recall from 1,000 simulations).}
    \label{fig_02}
\end{figure}

When retrieving one symptom at a time, we used a strict variance threshold of 80\%. The median True Positive Incidence Rate (TPIR) across all 80 symptoms was 85\% (range: 32\%--100\%). Seventy-six out of 80 symptoms had a TPIR above 50\% (see Fig.~\ref{fig_02}), confirming that the method consistently retrieves the simulated symptoms. 
As expected, TPIR decreases when symptoms are more similar to one another, due to a higher risk of confounding. Notably, there were two pairs of near-synonymous symptoms: delayed orgasm/unable to have orgasm, and hoarseness/voice quality.

Overall, the method demonstrated robust performance in recovering the most relevant PRO-CTCAE items from complex AE profiles, while limiting the size of the instrument.
\subsection{Selected Oncology Trials}

We processed two consecutive oncology trials to compare the performance of our automated selector with a manual symptom selection approach.

The first study, a phase I expansion trial (NCT02064387), included 35 patients with multiple myeloma who reported a total of 88 distinct adverse events. Using an information threshold of 90\%, our system identified an optimal set of 16 PRO-CTCAE symptoms (see Table~\ref{tab:pro_utility_table}). 
All selected symptoms had exact PT matches in the trial except for \emph{Mouth/throat sores}, where a related PT, \emph{Oropharyngeal pain}, was reported in three patients.
Overall, 24 PT events reported in the trial had an exact match to a PRO-CTCAE symptom. However, the diversity-weighted variance ratio cut-off excluded nine of these due to excessive redundancy. Additional terms could be included by raising the information threshold above 90\%.

We compared our list of 16 symptoms with the 15 symptoms that were manually selected for the subsequent phase II trial (NCT03525678). The two lists shared eight symptoms: \emph{Blurred vision}, \emph{Bruising}, \emph{Chills}, \emph{Decreased appetite}, \emph{Nosebleed}, \emph{Shortness of breath}, \emph{Watery eyes}, and \emph{Mouth/throat sores}.
Four symptoms chosen manually (\emph{Constipation}, \emph{Fatigue}, \emph{Nausea}, \emph{Vomiting}) were reported in the phase I trial but excluded by our method due to redundancy. The remaining three manually selected symptoms (\emph{General pain}, \emph{Heart palpitations}, \emph{Pain and swelling at injection site}) had relevance scores between 68\% and 75\%, below the threshold used in the automated selection.

In the phase II trial (NCT03525678), 176 distinct adverse events were reported among 221 patients, with 19 PTs mapping exactly to PRO-CTCAE symptoms.
The optimal and manual symptom lists achieved comparable AE retrieval performance in this study: our method identified 11 of its 16 selected symptoms as exact matches, while the manual list identified 11 of 15.

Overall, our findings indicate that the automated PRO-CTCAE symptom selection method provides practical value and performance comparable to a traditional manual approach. 
Importantly, it offers an objective, reproducible framework and quantitative justification for symptom selection.

\section{Discussion}

This work demonstrates a data-driven approach to designing PRO-CTCAE item sets. By adding a semantic embedding layer (Safeterm) on top of MedDRA, it is possible to construct objective metrics to solve such pratical problems numerically. As such, selection becomes objective and reproducible, with performance quantifiable with simulations. 

Our MedDRA semantic assessment of PRO-CTCAE symtoms revealed substantial overlap (see Figure~\ref{fig_02}) for several items such as voice quality, delayed orgasm, nausea/vomiting, and incontinence. Although the PRO-CTCAE was rigorously validated \citep{Dueck2015PROCTCAE}, these findings suggest that patients may have difficulty distinguishing certain closely related symptoms. From a MedDRA perspective, the current terminology also appears insufficiently granular to fully separate these concepts.

A limitation of our approach is the reliance on historical AE data. Sparse or unrepresentative profiles may lead to omission of emerging toxicities or inclusion of less clinically relevant terms. To mitigate this, the system allows users to curate and weight the input AE list. Future work may incorporate additional data sources or structured expert input to refine the historical toxicity profile further.

As the number of retained symptoms increases, patient burden and dropout risk increase as well. Our method returns a ranked list of symptoms prioritized for relevance and semantic diversity, with the recommended list size controlled through the cumulative information fraction and diversity score. Additional work may be necessary on these metrics to optimize balance between comprehensiveness and burden.

This spectral decomposition approach offers a deterministic and computationally efficient alternative to iterative DPP sampling. Whereas DPPs select diverse subsets by sampling from a probability distribution, the spectral method directly identifies the most informative diversity axes and ranks items by their leverage on these axes. This preserves the diversity-promoting properties of DPPs while avoiding sampling variability and providing explicit control over the retained dimensionality.

Finally, our utility function was constructed empirically to achieve a desired relevance saturation profile, with a weighting term to break ties among highly relevant symptoms. It performed well in both simulation and real-world evaluation, but its formulation is specific to the Safeterm embedding space and may require adjustment if the underlying model is updated. Future research could explore alternative functional forms or data-driven methods to learn optimal weighting.

\section{Conclusion}

We developed an automated pipeline for selecting an optimal set of PRO-CTCAE symptoms using frequency lists of MedDRA PTs from historical trials or expert opinion. Performance was validated through simulations and a retrospective case study. The method balances relevance to prior AEs with diversity among selected items, minimizing patient burden while maximizing signal coverage. This approach provides an objective, reproducible framework for PRO-CTCAE design, streamlining item selection in oncology trials.

%\section*{Statements and Declarations}

%\textbf{Funding:} No funding was provided for this study.\\[4pt]

%\textbf{Conflicts of interest:} 
%The authors are affiliated with ClinBAY Ltd; they declare no financial or commercial conflict beyond this.\\[4pt]

%\textbf{Author contributions:} 
%All authors contributed equally to concept, methodology, data analysis, and writing.\\[4pt]

%\textbf{Ethics approval:} Not applicable; retrospective analysis of existing data.\\[4pt]

\textbf{Data availability:}\\
\url{https://clinicaltrials.gov/study/NCT02064387}\\
\url{https://clinicaltrials.gov/study/NCT03525678}\\

\bibliographystyle{unsrtnat}
\bibliography{main}

\clearpage
\begin{longtable}{p{2.2cm} p{3.0cm} p{3.8cm} p{3.0cm}}
    \caption{MedDRA v28.1 Mapping of PRO-CTCAE Symptoms (v5/12/2025)}
    \label{tab:meddra_map} \\
    \hline
    \textbf{PRO-CTCAE Category} & \textbf{PRO-CTCAE Symptom} & \textbf{Preferred Term (MedDRA)} & \textbf{Lower Level Term (MedDRA)} \\
    \hline
    \endfirsthead
    
    \hline
    \textbf{PRO-CTCAE Category} & \textbf{PRO-CTCAE Symptom} & \textbf{Preferred Term (MedDRA)} & \textbf{Lower Level Term (MedDRA)} \\
    \hline
    \endhead
    
    \hline
    \multicolumn{4}{r}{\textit{Continued on next page}} \\
    \endfoot
    
    \hline
    \endlastfoot
    
    Oral & Dry mouth & Dry mouth & Dry mouth \\
         & Difficulty swallowing & Dysphagia & Swallowing difficult \\
         & Mouth/throat sores & Oropharyngeal pain & Sore throat \\
         &  & Oral pain & Sore mouth \\
         & Cracking at corners of mouth (cheilosis/cheilitis) & Cheilosis & Cheilosis \\
         &  & Cheilitis & Cheilitis \\
         & Voice quality changes & Dysphonia & Dysphonia \\
         & Hoarseness & Dysphonia & Hoarseness \\
    Respiratory & Shortness of breath & Dyspnoea & Shortness of breath \\
         & Cough & Cough & Cough \\
         & Wheezing & Wheezing & Wheezing \\
    Neurological & Dizziness & Dizziness & Dizziness \\
         & Numbness \& tingling & Hypoaesthesia & Numbness \\
         &  & Paraesthesia & Tingling \\
    Visual/Perceptual & Blurred vision & Vision blurred & Blurred vision \\
         & Flashing lights & Photopsia & Flashing lights \\
         & Visual floaters & Vitreous floaters & Vitreous floaters \\
         & Watery eyes & Lacrimation increased & Epiphora \\
         & Ringing in ears & Tinnitus & Ringing in ears \\
    Sleep/Wake & Insomnia & Insomnia & Insomnia \\
         & Fatigue & Fatigue & Fatigue \\
    Sexual & Achieve/maintain erection & Erectile dysfunction & Erection failure \\
         & Ejaculation & Ejaculation disorder & Ejaculation disorder \\
         & Decreased libido & Libido decreased & Hypoactive sexual desire disorder \\
         & Delayed orgasm & Orgasm abnormal & Delayed orgasm \\
         & Unable to have orgasm & Anorgasmia & Anorgasmia \\
         & Pain w/ sexual intercourse & Dyspareunia & Dyspareunia \\
    Cardio/Circulatory & Swelling & Swelling & Swelling \\
         & Heart palpitations & Palpitations & Palpitations \\
    Cutaneous & Rash & Rash & Rash \\
         & Skin dryness & Dry skin & Dry skin \\
         & Acne & Acne & Acne \\
         & Hair loss & Alopecia & Hair loss \\
         & Itching & Pruritus & Itching \\
         & Hives & Urticaria & Hives \\
         & Hand-foot syndrome & Palmar-plantar erythrodysaesthesia syndrome & Hand and foot syndrome \\
         & Nail loss & Onychomadesis & Nail loss \\
         & Nail ridging & Nail ridging & Nail ridging \\
         & Nail discoloration & Nail discolouration & Nail discoloration \\
         & Sensitivity to sunlight & Photosensitivity reaction & Photosensitivity \\
         & Bed/pressure sores & Decubitus ulcer & Decubitus ulcer \\
         & Radiation skin reaction & Radiation skin injury & Dermatitis radiation \\
         & Skin darkening & Skin discolouration & Darkened skin \\
         & Stretch marks & Skin striae & Stretch marks \\
    Gastrointestinal & Taste changes & Dysgeusia & Dysgeusia \\
         & Decreased appetite & Decreased appetite & Decreased appetite \\
         & Nausea & Nausea & Nausea \\
         & Vomiting & Vomiting & Vomiting \\
         & Heartburn & Dyspepsia & Heartburn \\
         & Gas & Flatulence & Gas \\
         & Bloating & Abdominal distension & Bloating \\
         & Hiccups & Hiccups & Hiccups \\
         & Constipation & Constipation & Constipation \\
         & Diarrhea & Diarrhoea & Diarrhea \\
         & Abdominal pain & Abdominal pain & Abdominal pain \\
         & Fecal incontinence & Anal incontinence & Fecal incontinence \\
    Attention/Memory & Concentration & Disturbance in attention & Concentration impaired \\
         & Memory & Amnesia & Loss of memory \\
    Mood & Anxious & Anxiety & Anxiety \\
         & Discouraged & Discouragement & Discouragement \\
         & Sad & Depressed mood & Feeling sad \\
    Pain & General pain & Pain & Pain \\
         & Headache & Headache & Headache \\
         & Muscle pain & Myalgia & Muscle pain \\
         & Joint pain & Arthralgia & Joint pain \\
    Genitourinary & Irregular periods/vaginal bleeding & Menstruation irregular & Irregular menstruation \\
         & Missed expected menstrual period & Menstruation delayed & Menstruation delayed \\
         & Vaginal discharge & Vaginal discharge & Vaginal discharge \\
         & Vaginal dryness & Vulvovaginal dryness & Vaginal dryness \\
         & Painful urination & Dysuria & Painful urination \\
         & Urinary urgency & Micturition urgency & Urinary urgency \\
         & Urinary frequency & Pollakiuria & Urinary frequency \\
         & Change in usual urine color & Chromaturia & Discoloration urine \\
         & Urinary incontinence & Urinary incontinence & Urinary incontinence \\
    Miscellaneous & Breast swelling/tenderness & Breast tenderness & Breast tenderness \\
         & & Breast swelling & Breast swelling \\
         & Bruising & Contusion & Bruising \\
         & Chills & Chills & Chills \\
         & Increased sweating & Hyperhidrosis & Hyperhidrosis \\
         & Decreased sweating & Hypohidrosis & Hypohidrosis \\
         & Hot flashes & Hot flush & Hot flashes \\
         & Nosebleed & Epistaxis & Nosebleed \\
         & Pain/swelling at injection site & Injection site pain & Injection site pain \\
         & & Injection site swelling & Injection site swelling \\
         & Body odor & Skin odour abnormal & Body odor \\
\end{longtable}

\clearpage	

\begin{table}[ht!]
    \centering
    \caption{Optimal PRO-CTCAE Symptoms based on NCT02064387 multiple myeloma expansion phase trial (N=35), including relevance, utility metrics, and mapped adverse events for each PRO symptom.}
    \label{tab:pro_utility_table}
    \small
    \setlength{\tabcolsep}{6pt}
    \renewcommand{\arraystretch}{1.25}

    \begin{tabular}{ccccc l}
    \toprule
    \textbf{Relevance} & \textbf{Weight} & \textbf{Utility} &
    \textbf{Diversity Leverage} & \textbf{PRO Symptom} & \textbf{Most Relevant AEs [\# affected]} \\
    \midrule

    1.000 & 0.050 & 0.987 & 0.841 & Chills &
    Chills [9] \\

    1.000 & 0.011 & 0.983 & 0.822 & Hair loss &
    Alopecia [2] \\

    1.000 & 0.028 & 0.985 & 0.760 & Decreased appetite &
    Decreased appetite [5] \\

    1.000 & 0.077 & 0.990 & 0.758 & Cough &
    Cough [14] \\

    0.932 & 0.017 & 0.935 & 0.750 & Mouth/throat sores &
    Oropharyngeal pain [3] \\

    1.000 & 0.022 & 0.984 & 0.746 & Watery eyes &
    Lacrimation increased [4] \\

    1.000 & 0.099 & 0.992 & 0.724 & Blurred vision &
    Vision blurred [18] \\

    1.000 & 0.028 & 0.985 & 0.704 & Bruising &
    Contusion [5] \\

    1.000 & 0.017 & 0.984 & 0.664 & Insomnia &
    Insomnia [3] \\

    1.000 & 0.055 & 0.988 & 0.647 & Diarrhea &
    Diarrhoea [10] \\

    1.000 & 0.017 & 0.984 & 0.630 & Nosebleed &
    Epistaxis [3] \\

    1.000 & 0.022 & 0.984 & 0.622 & Rash &
    Rash [4] \\

    1.000 & 0.011 & 0.983 & 0.597 & Dizziness &
    Dizziness [2] \\

    1.000 & 0.011 & 0.983 & 0.593 & Urinary urgency &
    Micturition urgency [2] \\

    1.000 & 0.033 & 0.985 & 0.580 & Shortness of breath &
    Dyspnoea [6] \\

    1.000 & 0.028 & 0.985 & 0.552 & Joint pain &
    Arthralgia [5] \\

    \bottomrule
    \end{tabular}

\end{table}

\clearpage
\begin{table}[ht]
	\centering
	\caption{Summary statistics for PRO-CTCAE selection performance across simulations.}
	\label{tab:simulation_summary}
	\begin{tabular}{lcccc}
	\hline
	 & \textbf{Size Simulated} & \textbf{Recall} & \textbf{Precision} & \textbf{F1} \\
	\hline
	\textbf{Mean} & 22.50 & 0.70 & 0.72 & 0.70 \\
	\textbf{Std} & 10.37 & 0.12 & 0.12 & 0.08 \\
	\textbf{Min} & 5.00 & 0.31 & 0.18 & 0.25 \\
	\textbf{Median} & 23.00 & 0.67 & 0.74 & 0.70 \\
	\textbf{Max} & 40.00 & 1.00 & 1.00 & 1.00 \\
	\hline
	\end{tabular}
	\end{table}

\end{document}